\documentclass[10pt,twocolumn,letterpaper]{article}

\usepackage{ijcb}
\usepackage{times}
\usepackage{epsfig}
\usepackage{graphicx}
\usepackage{amsmath}
\usepackage{amssymb}
\usepackage{makecell}
\usepackage{multirow}
\usepackage{tabularx}
\usepackage[final]{microtype}
\usepackage{booktabs}
\usepackage[accsupp]{axessibility} 

\usepackage[pagebackref=true,breaklinks=true,colorlinks,bookmarks=false]{hyperref}

\ijcbfinalcopy 

\ifijcbfinal\fi
\begin{document}

\title{Human Identification at a Distance: Challenges, Methods and Results \\ on the Competition HID 2025\vspace{-1.5em}}

\author{
Jingzhe Ma$^{1}$, Meng Zhang$^{2}$, Jianlong Yu$^{3}$, Kun Liu$^{4}$, Zunxiao Xu$^{5}$, Xue Cheng$^{6}$, Junjie Zhou$^{7}$, \\ 
Yanfei Wang$^{8}$, Jiahang Li$^{8}$, Zepeng Wang$^{9}$, Kazuki Osamura$^{10}$, Rujie Liu$^{2}$, Narishige Abe$^{10}$, \\
Jingjie Wang$^{3}$, Shunli Zhang$^{3}$, Haojun Xie$^{4}$, Jiajun Wu$^{4}$, Weiming Wu$^{4}$, Wenxiong Kang$^{4}$, \\
Qingshuo Gao$^{5}$, Jiaming Xiong$^{5}$, Xianye Ben$^{5}$, Lei Chen$^{5}$, Lichen Song$^{6}$, Junjian Cui$^{6}$, \\
Haijun Xiong$^{11}$, Junhao Lu$^{12}$, Bin Feng$^{11}$, Mengyuan Liu$^{13}$, Ji Zhou$^{8}$, Baoquan Zhao$^{8}$, Ke Xu$^{9}$, \\
Yongzhen Huang$^{14,15}$, Liang Wang$^{16}$, Manuel J Marin-Jimenez$^{17}$, Md Atiqur Rahman Ahad$^{18}$, Shiqi Yu$^{*}$$^{19}$ \\
{\small
$^{1}$Shenzhen Polytechnic University, China. 
$^{2}$Fujitsu Research and Development Center Company Ltd., China.}\\
{\small
$^{3}$Beijing Jiaotong University, China.
$^{4}$South China University of Technology, China. 
$^{5}$Shandong University, China. 
} \\
{\small
$^{6}$EVERSPRY, China. 
$^{7}$Wuhan University of Technology, China.
$^{8}$Sun Yat-sen University, China. 
} \\
{\small
$^{9}$Shanghai Jiao Tong University, China.
$^{10}$Fujitsu Ltd., Japan.
$^{11}$Huazhong University of Science and Technology, China.
}\\
{\small
$^{12}$Hefei University of Technology, China.
$^{13}$Peking University, Shenzhen Graduate School, China.
$^{14}$Beijing Normal University, China.
}\\
{\small
$^{15}$Watrix Technology Limited Co. Ltd., China.
$^{16}$Institute of Automation, Chinese Academy of Sciences, China.
}\\
{\small
$^{17}$University of Cordoba, Spain.
$^{18}$University of East London, UK.
$^{19}$Southern University of Science and Technology, China.
}\\
{\small {\tt \url{https://hid.iapr-tc4.org/}} $^{*}$ Corresponding author: Shiqi Yu, yusq@sustech.edu.cn}
\vspace{-1.5em}}
\maketitle
\thispagestyle{empty}

\begin{abstract}
\vspace{-0.5em}
Human identification at a distance (HID) faces challenges due to the difficulty of acquiring traditional biometric modalities like face and fingerprints. Gait recognition offers a viable solution since it can be captured at a distance.
To promote progress in gait recognition and provide a fair evaluation platform, the International Competition on Human Identification at a Distance (HID) has been organized annually since 2020. 
Since 2023, the competition has adopted the challenging SUSTech-Competition dataset, which includes significant variations in clothing, carried objects, and view angles. 
No training data is provided, requiring participants to train their models using external datasets. 
Each year, the competition applies a different random seed to generate distinct evaluation splits, reducing the risk of overfitting and ensuring fair evaluation of cross-domain generalization. 
Although the previous two competitions (HID 2023 and HID 2024) already utilized this dataset, HID 2025 aimed explicitly to explore whether algorithmic improvements could surpass the accuracy limits observed previously. 
Despite these heightened challenges, participants again demonstrated significant advancements, with the highest accuracy reaching 94.2\%, setting a new benchmark for this dataset.
We also analyze key technical trends and outline potential directions for future research on gait recognition.
\end{abstract}

\vspace{-1.5em}
\section{Introduction}
\vspace{-0.5em}
Human identification at a distance (HID) is crucial for public security and surveillance because, in long-range monitoring and crowded environments, it is often difficult to acquire clear facial images or other conventional biometric data~\cite{problem2005}. Traditional methods such as face or fingerprint recognition are less effective under these conditions. As a result, developing reliable techniques that can identify individuals from afar has become an essential research focus. Among various distance-based biometric modalities, gait recognition, which identifies people by their unique walking patterns, emerges as a particularly promising solution because it can operate effectively under these challenging conditions~\cite{connor2018, nixon2006book}.

Recent advances in deep learning have markedly improved gait recognition accuracy. Modern convolutional neural network architectures (e.g. GaitSet~\cite{gaitset}, GaitBase~\cite{fan2023exploring}, and DeepGaitv2~\cite{fan2023exploring}, etc.) now achieve very high performance on controlled gait benchmarks. Also, some recent methods like BigGait~\cite{biggait} and DenosingGait~\cite{jin2025denoising} employs large vision models to improve gait recognition. However, performance remains sensitive to environmental and appearance variations. Empirical studies show that models trained on standard laboratory datasets often suffer substantial accuracy drops when tested on unconstrained “in‑the‑wild” data~\cite{fan2023exploring}. Factors such as changes in clothing, carried objects, camera viewpoint, or scene occlusions can significantly degrade the gait representation. More generally, domain shifts between training and deployment data severely limit cross-dataset generalization. Thus, despite impressive benchmarks in ideal settings, gait recognition systems still struggle to deliver robust accuracy and generality in realistic deployments.

To support progress under practical conditions, the International Competition on Human Identification at a Distance (HID) has served as a benchmark platform since 2020. HID 2025 is the sixth edition of this competition. In the first three editions (2020 to 2022), the challenge relied on a subset of the CASIA-E dataset~\cite{casiae}, where recognition accuracy improved rapidly and eventually reached 95.9\% by 2022. To make the task more challenging, HID 2023 adopted the \textbf{SUSTech-Competition} dataset, which includes large variations in clothing, carried items, and camera viewpoints. At the same time, the organizers stopped providing a fixed training set, requiring participants to train models using external data. This change introduced significant domain gaps between training and test data.

\begin{figure*}[t]
\begin{center}
\vspace{-1.0em}
\includegraphics[width=\linewidth]{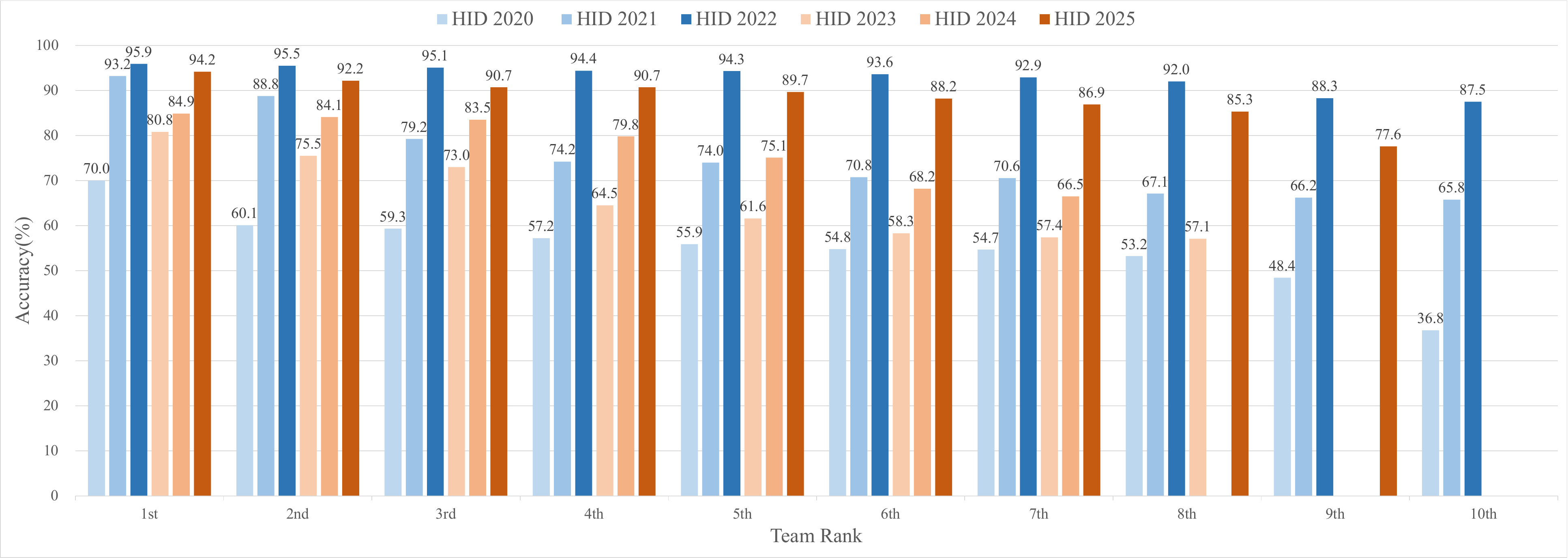}
\end{center}
\vspace{-1.0em}
\caption{Performance comparison of the top-10 final ranks across the last six HID competitions. The x-axis represents the final rank (e.g., 1st, 2nd). This visualization highlights the performance trends for each rank over the years. The orange-based bars show results for HID 2025 on the challenging SUSTech-Competition dataset, while the blue-based bars represent results on the CASIA-E dataset used from 2020-2022.}
\vspace{-1.0em}
\label{fig:results-2020-2025}
\end{figure*}

In HID 2025, the \textbf{SUSTech-Competition} dataset is used again, but with a new split of the data to prevent overfitting. The \textbf{primary motivation of HID 2025} is to explore whether recent algorithmic advances could push performance beyond the limits observed in 2023 to 2024.  In practice, participants did achieve notable progress: several teams exceeded 90\% rank-1 accuracy, and the best performing entry attained a record 94.2\% accuracy on this challenging benchmark. These results set a new standard and demonstrate continued improvement in robust gait recognition under stringent conditions. To provide broader context, Figure~\ref{fig:results-2020-2025} illustrates the overall performance trend for each of the top-10 ranks over the past six years, showing the evolution of state-of-the-art accuracy at different competitive tiers.

This paper is prepared by the competition organizers together with participants from the top nine teams. It provides a summary of the HID 2025 competition. Section 2 describes the competition setup, the dataset, and the evaluation procedures. Section 3 presents the results and the main methods developed by the top nine teams. Section 4 analyzes the technical approaches and trends observed in this edition. Section 5 concludes the paper and outlines possible directions for future research.

\vspace{-0.5em}
\section{Dataset and Competition Details}
\vspace{-0.5em}

\subsection{Dataset}
\vspace{-0.5em}
HID 2025 utilizes the SUSTech-Competition gait dataset, identical to the one introduced in the HID 2023 and 2024. 
However, for this year’s challenge the test set was defined by applying a new random split, ensuring that none of the test sequences overlap with those used in prior competitions. 

\begin{figure}[htbp]
\begin{center}
\vspace{-1.0em}
\includegraphics[width=0.38\linewidth]{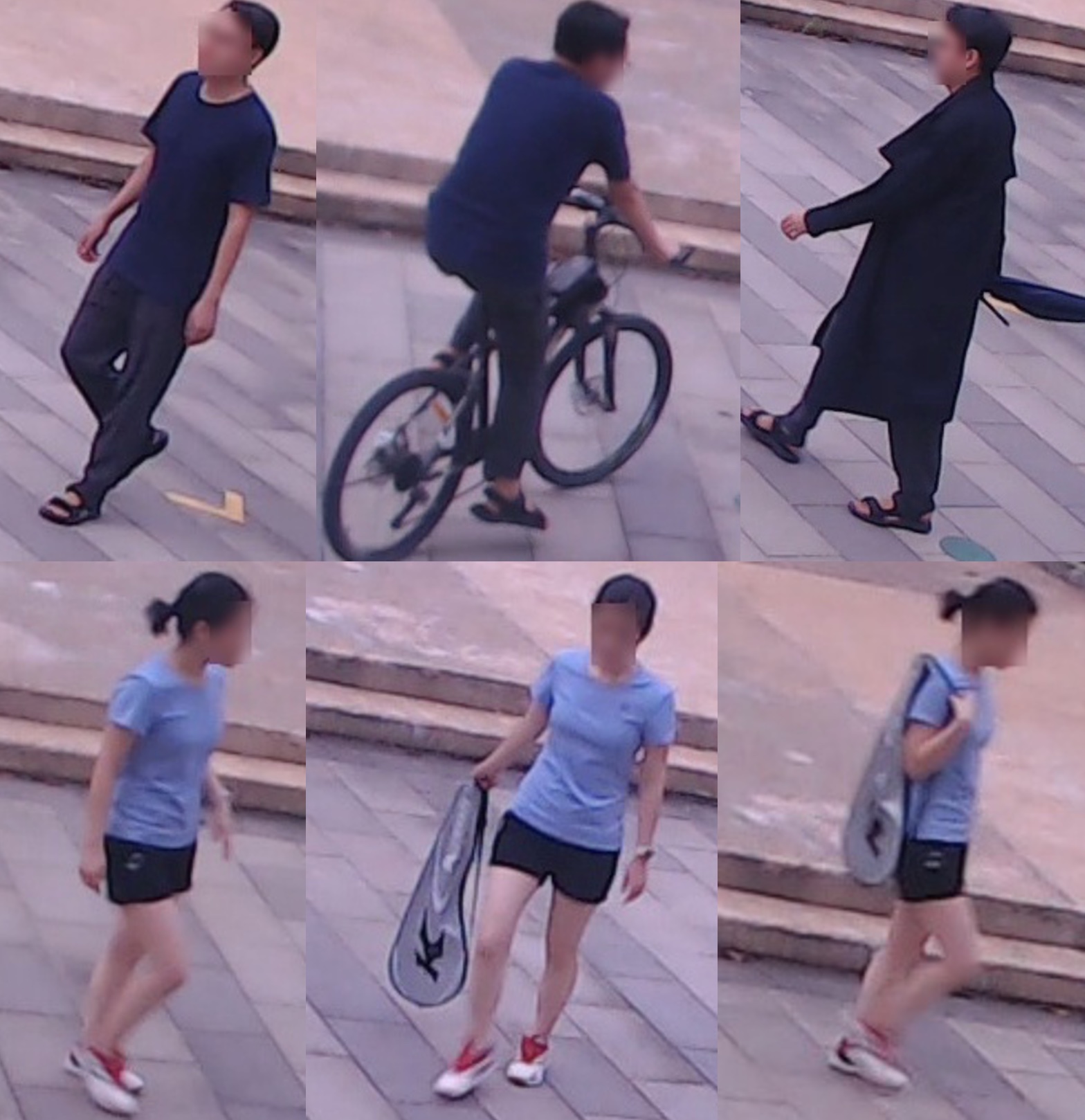}
\includegraphics[width=0.57\linewidth]{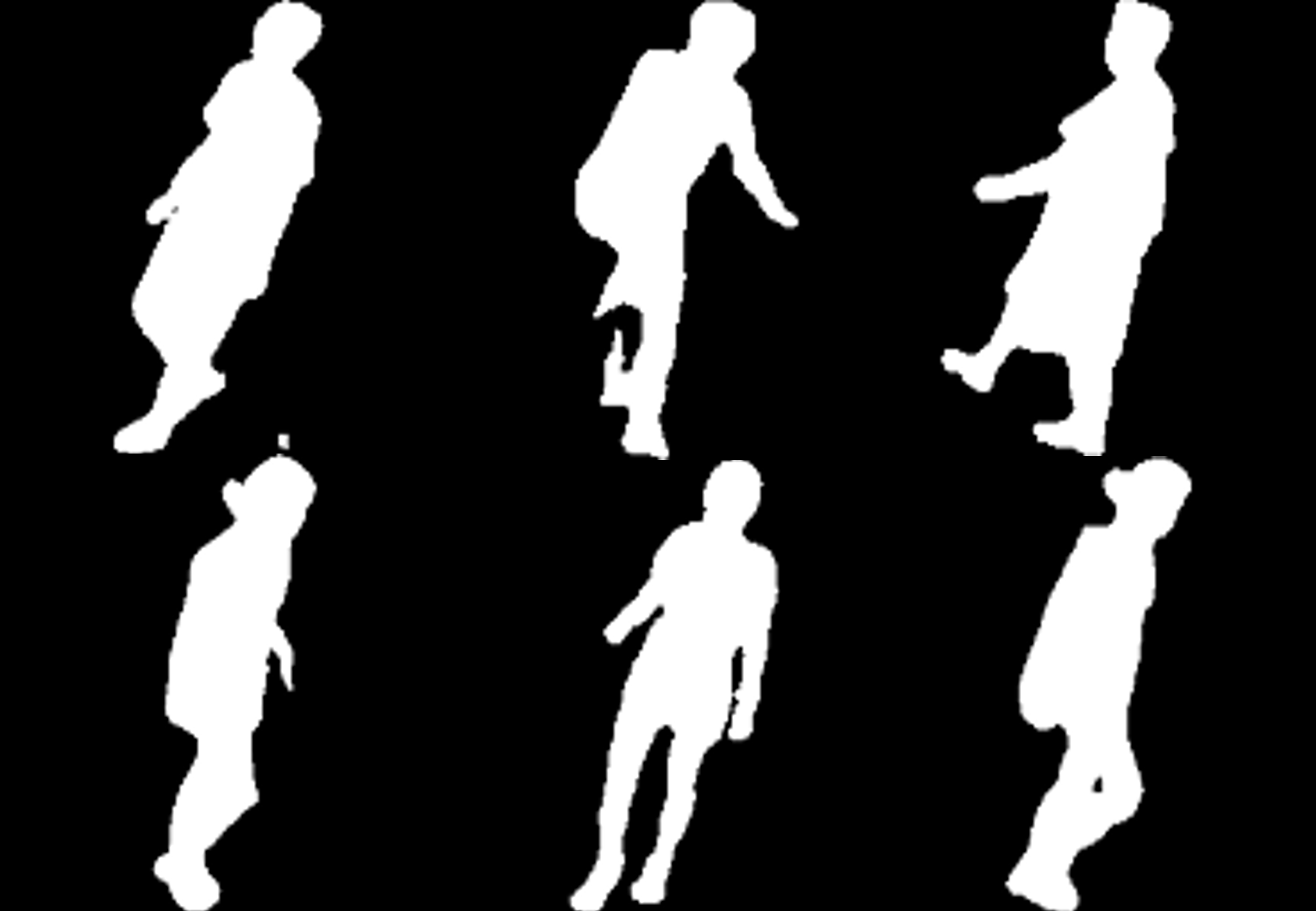}
\end{center}
\vspace{-1.0em}
\caption{Some RGB images and their corresponding silhouettes from the dataset \textit{SUSTech-Competition}. Many variations are included in the dataset.}
\vspace{-2.0em}
\label{fig:dataset}
\end{figure}

The dataset was collected during the summer of 2022, with the approval of the Southern University of Science and Technology Institutional Review Board. The complete dataset comprises 859 subjects and encompasses various variations, including clothing, carrying conditions, and view angles, as shown in Figure \ref{fig:dataset}. To alleviate the participants' data preprocessing burden, we provided human body silhouettes in the competition. These silhouettes were obtained from the original videos using a deep person detector and a segmentation model provided by our sponsor, Watrix Technology.

All silhouette images were resized to a fixed size of $128 \times 128$, as illustrated in Figure \ref{fig:dataset}. We intentionally did not manually remove low-quality silhouettes, as the presence of noise reflects real-world application scenarios and adds to the challenge of the competition. This approach ensures that the competition provides a realistic simulation of real applications.

The same as HID 2023 and HID 2024, we did not provide a specific training set to participants. Instead, participants can use any dataset, such as CASIA-B~\cite{casiab}, OU-MVLP~\cite{dataset2017OUMVLP}, CASIA-E~\cite{casiae}, GREW~\cite{grew_iccv2021}, Gait3D~\cite{gait3d_cvpr2022}, SUSTech-1K~\cite{lidargait}, CCPG~\cite{ccpg}, CCGR~\cite{ccgr}, DroneGait~\cite{drone24} or their own datasets, to train their models. Since the training set and the test set would be from different datasets, the cross-domain challenge will be introduced. Participants have to consider this aspect to achieve good results. The gallery in the test set consists of only one sequence per subject, with the labels of the sequences provided to the participants. The probe set contains only a single randomly selected sequence for each subject. The probe samples may exhibit variations in view, clothing, carrying conditions, and occlusions compared to the gallery samples. These settings make the competition challenging and align it closely with real applications.

Specifically, although the test probe set for HID 2023, HID 2024, and HID 2025 are all randomly sampled from the \textit{SUSTech-Competition} dataset, the test set used in HID 2025 is different from those of HID 2023 and HID 2024 due to the use of different random seeds. Furthermore, all samples from the probe sets of HID 2023 and HID 2024, as well as the probe set provided for HID 2025, are strictly prohibited from being used in any manner during the training phase.

\vspace{-0.5em}
\subsection{Performance metric}
\vspace{-0.5em}
Similar to the previous competitions in the series, the evaluation metric is considered as the rank-1 \textit{Accuracy}. It provides a straightforward and easily implemented metric. It can be calculated as follows: $\text{Accuracy} = \frac{TP}{N_{probe}}$,
where $TP$ is the number of probe samples that are correctly identified, and $N_{probe}$ means the total number of probe samples.

\vspace{-0.5em}
\subsection{Competition policies}
\vspace{-0.5em}

The evaluation process for HID 2025 was further refined to enhance both fairness and the assessment of model robustness, while maintaining user convenience and security against hacking attempts. The main policies are as follows. (1) The competition comprises two phases. Phase 1 (Feb 28 to May 6, 2025) evaluates the models on a randomly selected 10\% subset of the test set. Phase 2 (May 7 to May 16, 2025) uses a different random subset covering the entire test set, and its results are considered final. This two-stage evaluation mitigates label leakage and enables a more reliable assessment of model robustness. (2) To prevent the ID labels of the probe set from being deduced through multiple submissions, each team is limited to a maximum of 5 submissions per day during the first phase and 2 submissions per day during the second phase. Only one CodaLab ID is allowed per team, and only registrations using institutional emails (not public emails) are accepted. (3) The accuracy of the submissions is automatically evaluated on CodaLab, and the rankings are updated on the scoreboard accordingly. This immediate feedback ensures a user-friendly evaluation process. (4) The top teams on the final scoreboard are required to submit their codes to the organizers. The submitted codes are executed to reproduce their results, and the reproduced results should align with those displayed on the CodaLab scoreboard. 

\vspace{-0.5em}
\subsection{Competition statistics}
\vspace{-0.5em}

\begin{figure*}[htbp]
\begin{center}
\vspace{-1.0em}
\includegraphics[width=0.75\linewidth]{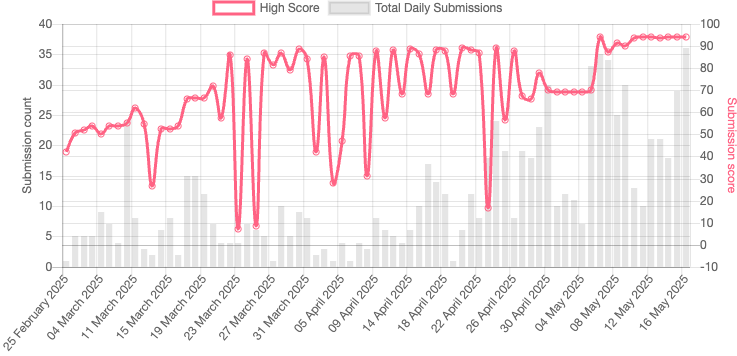}
\vspace{-1.0em}
\caption{The best scores and the numbers of submissions of each day during the competition.}
\vspace{-1.0em}
\label{fig:statis}
\end{center}
\end{figure*}

A total of 103 registrations were received for HID 2025, and registrations with public emails (e.g., Gmail) had been rejected. Among the valid registrations, which amounted to 23 teams submitted their results to CodaLab during the second phase. The best scores and the numbers of submissions of each day can be found in Figure~\ref{fig:statis}. Most of the teams actively participated in the competition. The programs of the top teams were carefully evaluated to verify the reproducibility of their results. After a thorough evaluation, the top 9 teams were selected based on their performance. The methods employed by these top teams will be introduced in the following section.

\vspace{-0.5em}
\section{Methods of the Top Teams}
\vspace{-0.5em}

\begin{table*}[t]
\centering
\vspace{-1.0em}
\caption{The technologies used by the top 9 teams and their accuracies in HID 2025.}
\label{tab:top2025}
\resizebox{\textwidth}{!}{%
\begin{tabular}{@{}c|c|c|c|c|c|c|c|c|c@{}}
\toprule
Team rank         & \textbf{1}                                                                                                     & \textbf{2}                                                      & \textbf{3}                                                       & \textbf{4}                                                                        & \textbf{5}                                                     & \textbf{6}                                                 & \textbf{7}                                                                        & \textbf{8}                                        & \textbf{9}                                                                  \\ \midrule
Team Name         & BRAVO-FJ                                                                                                       & BJTU-SSLL                                                       & SCUT-BIPLAB                                                      & :)                                                                                & league                                                         & HUST-MCLAB                                                 & ITZY                                                                              & sysu                                              & SJTU-ICL                                                                    \\ \midrule
Data cleaning     & $\checkmark$                                                                                                   & $\times$                                                        & $\checkmark$                                                     & $\times$                                                                          & $\times$                                                       & $\times$                                                   & $\times$                                                                          & $\times$                                          & $\times$                                                                    \\ \midrule
Data alignment    & $\checkmark$                                                                                                   & $\checkmark$                                                    & $\checkmark$                                                     & $\checkmark$                                                                      & $\checkmark$                                                   & $\checkmark$                                               & $\checkmark$                                                                      & $\checkmark$                                      & $\checkmark$                                                                \\ \midrule
Data augmentation & $\checkmark$                                                                                                   & $\checkmark$                                                    & $\checkmark$                                                     & $\checkmark$                                                                      & $\checkmark$                                                   & $\checkmark$                                               & $\checkmark$                                                                      & $\checkmark$                                      & $\checkmark$                                                                \\ \midrule
Re-ranking        & $\checkmark$                                                                                                   & $\checkmark$                                                    & $\checkmark$                                                     & $\checkmark$                                                                      & $\checkmark$                                                   & $\checkmark$                                               & $\checkmark$                                                                      & $\checkmark$                                      & $\checkmark$                                                                \\ \midrule
Ensemble          & $\checkmark$                                                                                                   & $\checkmark$                                                    & $\checkmark$                                                     & $\checkmark$                                                                      & $\checkmark$                                                   & $\checkmark$                                               & $\checkmark$                                                                      & $\checkmark$                                      & $\checkmark$                                                                \\ \midrule
Training data     & \makecell{CCPG, HID2022, \\ CASIA-B,  Gait3D, \\ SUSTech-1K, \\ 1006 internal dataset, \\HID 2025 gallery set} & \makecell{CASIA-E, \\ HID 2025 gallery set}                     & \makecell{Gait3D,  CCGR-mini, \\ CASIA-E, OU-MVLP, \\ SUSTech-1K} & \makecell{CASIA-B,  OU-MVLP, \\Gait3D,  CASIA-E, \\CCPG, CCGR-mini, \\SUSTech-1K} & \makecell{Gait3D, HID2022}                                     & \makecell{SUSTech-1K, \\ Gait3D, HID 2022}                 & \makecell{CASIA-B,  OU-MVLP, \\Gait3D,  CASIA-E, \\CCPG,  HID 2022, \\SUSTech-1K} & \makecell{CASIA-B,  Gait3D, \\ CCPG, SUSTech-1K}  & \makecell{CASIA-B,  OU-MVLP, \\Gait3D,  CASIA-E, \\CCPG, GREW,\\SUSTech-1K} \\ \midrule
Pseudo-labelling  & \makecell{$\checkmark$ \\ (on HID2025  \\ gallery 859 samples)}                                                & \makecell{$\checkmark$ \\ (on HID2025  \\ gallery 859 samples)} & $\times$                                                         & $\times$                                                                          & \makecell{$\checkmark$ \\ (on HID2022 samples)}                & \makecell{$\checkmark$ \\ (on HID2022 samples)}            & \makecell{$\checkmark$ \\ (on HID2022 samples)}                                   & $\times$                                          & $\times$                                                                    \\ \midrule
Architecture      & \makecell{DeepGaitV2 P3D~\cite{fan2023exploring}}                                                              & \makecell{DeepGaitV2 3D~\cite{fan2023exploring}}                & \makecell{DeepGaitV2 P3D \& 3D~\cite{fan2023exploring}}          & \makecell{DeepGaitV2 P3D~\cite{fan2023exploring}}                                 & \makecell{DeepGaitV2 P3D, \\ SwinGait~\cite{fan2023exploring}} & \makecell{DeepGaitV2 P3D, \\ GLGait~\cite{peng2024glgait}} & \makecell{DeepGaitV2 P3D~\cite{fan2023exploring}}                                 & \makecell{DeepGaitV2 P3D~\cite{fan2023exploring}} & \makecell{DeepGaitV2 P3D, \\ SwinGait~\cite{fan2023exploring}}              \\ \midrule
Accuracy(\%)      & 94.2                                                                                                           & 92.2                                                            & 90.7 (90.687)                                                    & 90.7 (90.664)                                                                     & 89.7                                                           & 88.2                                                       & 86.9                                                                              & 85.3                                              & 77.6                                                                        \\ \bottomrule
\end{tabular}
}
\end{table*}

The competition organizers invited all top teams to submit their source code for review. Nine teams submitted their course code, and passed the code review.  The subsequent part of the section provides an in-depth exploration of the methods employed by each team. The technologies utilized by these teams, along with their corresponding results, are summarized in Table~\ref{tab:top2025}. 

\vspace{-0.5em}
\subsection{Team: BRAVO-FJ}
\vspace{-0.5em}

\noindent \textbf{Members:} Meng Zhang, Kazuki Osamura, Rujie Liu, and Narishige Abe \\
\noindent \textbf{Institutions:} Fujitsu Research and Development Center Company Ltd., and Fujitsu Ltd. \\
\noindent{\small \texttt{\{zhangmeng, osamura.kazuki, rjliu, abe.narishige\}@fujitsu.com}}

\begin{figure}[htbp]
	\centering
    \vspace{-1.em}
	\includegraphics[width=0.8\linewidth]{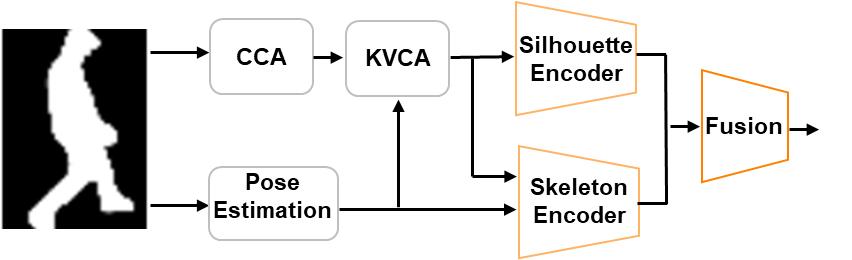}
	\caption{The framework of Team BRAVO-FJ's method.}
    \vspace{-2.em}
	\label{fig:team1}
\end{figure}

\noindent \textbf{Method:}
The method introduces a two-stage data cleaning process and a hybrid silhouette- and skeleton-based gait recognition model, as depicted in Figure~\ref{fig:team1}. Cleaning is done using Connected Component Analysis (CCA) and Keypoint Visibility-Confidence Analysis (KVCA). Samples with fragmented silhouettes or low-confidence keypoints are identified as low-quality and removed. CCA eliminates structural noise and corrupted regions, while KVCA filters remaining noisy frames. The cleaned silhouette sequences are then passed to a silhouette-based model, which is fused with a skeleton-based model for final recognition.

The silhouette model uses the DeepGaitV2 framework with a Pseudo 3D (P3D) convolution module. Five public datasets—CCPG, HID2022, SUSTech1K, Gait3D, and CASIA-B—are merged to create a training set covering scenarios like normal walking, object carrying, and varied clothing. With this setup, the model achieves 93.8\% accuracy. Adding an internal dataset of 1,006 subjects, each with 72 videos under 12 cameras and 3 clothing types, further improves accuracy to 94.2\%. Fine-tuning on the HID2025 gallery set enhances cross-domain performance. Data augmentation includes random perspective transformation, horizontal flipping, and rotation. Several optimization strategies are used during inference. A re-ranking step refines initial predictions. A two-step voting mechanism combines outputs from different training checkpoints and model variants to increase stability and accuracy.

\vspace{-0.5em}
\subsection{Team: BJTU-SSLL}
\vspace{-0.5em}

\noindent \textbf{Members:} Jianlong Yu and Jingjie Wang \\
\noindent \textbf{Supervisor:} Shunli Zhang \\
\noindent \textbf{Institutions:} Beijing Jiaotong University \\
\noindent{\small \texttt{\{24121438, 23111492, slzhang\}@bjtu.edu.cn}}

\begin{figure}[htbp]
	\centering
    \vspace{-1.0em}
	\includegraphics[width=\linewidth]{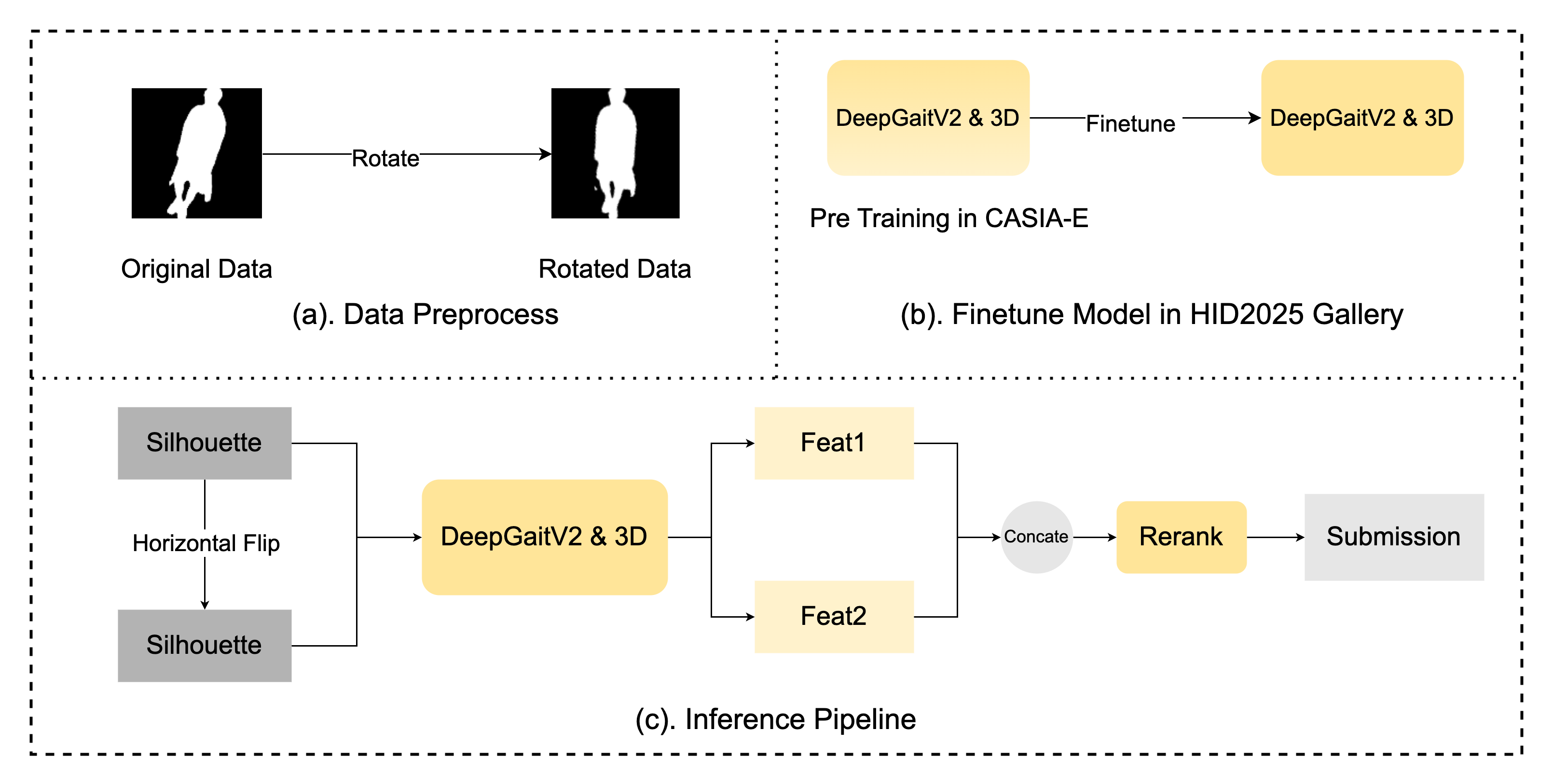}
    \vspace{-1.0em}
	\caption{The framework of Team BJTU-SSLL's method. }
    \vspace{-1.0em}
	\label{fig:team2}
\end{figure}

\noindent \textbf{Method:} 
The method is shown in Figure~\ref{fig:team2}, and it comprises three core components: dataset preprocessing, pre-training and fine-tuning, and inference.

\noindent\textbf{Dataset preprocessing.} To reduce noise interference in the HID2025 dataset, a rotation-based alignment is applied, as depicted in Figure~\ref{fig:team2}~(a). The method detects the midpoints of the upper and lower body on each silhouette contour, connects them to form a vertical axis, and rotates the image to align this axis with the x-axis.

\noindent\textbf{Pre-training and fine-tuning.} The DeepGaitV2-3D model~\cite{fan2023exploring} is first pre-trained on the CASIA-E dataset using random horizontal flipping and rotation, each with a probability of 0.2. 
To mitigate the domain gap between the source and target domains, they first augment the HID 2025 gallery dataset (859 sequences) using horizontal flipping and random rotation, expanding each sequence threefold. They then created a mixed dataset by combining this augmented gallery dataset with multiple sequences from 50 randomly selected identities from CASIA-E. This strategy aimed to maintain the previously learned feature distributions and avoid potential disruption caused by fine-tuning exclusively on the gallery data. Finally, the shallow feature extraction layers of the DeepGaitV2-3D model were fine-tuned using this mixed dataset, as depicted in Figure~\ref{fig:team2}~(b).

\noindent\textbf{Inference.} The inference process is illustrated in Figure~\ref{fig:team2}~(c). Each sample is first processed in its original form to extract features. The sample is then horizontally flipped and reprocessed. Features from the original and flipped samples are concatenated. A re-ranking step is applied to the combined features to generate the final predictions.

\noindent\textbf{Experimental setup.} Experiments are conducted on four RTX 3090 GPUs. The training involves a batch size of $8 \times 8$, with 200{,}000 iterations for pretraining and 10{,}000 for fine-tuning. All other hyperparameters follow the standard GaitGL settings defined in the OpenGait framework.

\vspace{-0.5em}
\subsection{Team: SCUT-BIPLAB}
\vspace{-0.5em}

\noindent \textbf{Members:} Kun Liu, Haojun Xie, Jiajun Wu, and Weiming Wu \\
\noindent \textbf{Supervisor:} Wenxiong Kang \\
\noindent \textbf{Institutions:} South China University of Technology \\
\noindent{\small \texttt{\{aulkun, 202130450186, 202130450162, auauweimingwu, auwxkang\}@mail.scut.edu.cn}}

\noindent \textbf{Method:} 
The training data is constructed by merging samples from five public gait datasets: OU-MVLP, Gait3D, SUSTech1K, CCGR-mini, and CASIA-E, resulting in a total of 11,469 unique identities. To enhance data quality and consistency, preprocessing steps include connected component filtering for noise removal, adaptive rotation correction based on silhouette slope estimation, resizing to $64 \times 64$, and optional center alignment. Both aligned and non-aligned versions are retained for model training and evaluation.

The backbone network adopts the DeepGaitV2 architecture, incorporating both 3D and P3D convolutional modules to capture temporal and spatial gait dynamics. A modification to the horizontal pyramid pooling module introduces bin numbers [16, 1]. On top of the standard Gait3D augmentation strategies, two additional techniques are employed: conventional random erasing and a custom variant termed “random erasing white,” which whitens random regions of the silhouette to simulate occlusions such as missing limbs or carried objects. Models are trained for 150,000 iterations with a batch size of $32 \times 4$.

During inference, predictions are obtained from both forward and reversed sequence inputs using the 3D and P3D models, applied to both aligned and non-aligned test sets. This results in eight predictions per subject. A majority voting strategy is used to determine the final identity. The P3D model alone achieves 89.732\% accuracy on the unaligned test set. Combining 3D and P3D models across both input directions raises accuracy to 90.617\%, while the full eight-way ensemble yields a final accuracy of 90.687\%.

\vspace{-0.5em}
\subsection{Team: :)}
\vspace{-0.5em}

\noindent \textbf{Members:} Zunxiao Xu, Qingshuo Gao, and Jiaming Xiong \\
\noindent \textbf{Supervisors:} Xianye Ben and Lei Chen \\
\noindent \textbf{Institutions:} Shandong University \\
\noindent{\small \texttt{\{zunxiaoxu, 202412757, 202100120251, benxianye, lei.chen\}@sdu.edu.cn}}

\noindent \textbf{Method:}
The training set is constructed by selecting identities from seven public gait datasets: CASIA-B, OU-MVLP, Gait3D, CASIA-E, CCPG, SUSTech1K, and CCGR-mini. All samples are aggregated into a unified dataset, known as TrainSet1. To improve spatial consistency across gait sequences, a PCA-based automatic alignment method is applied. For each sequence, gait contours are extracted from frames, the principal orientation is estimated via Principal Component Analysis (PCA), and a uniform rotation angle is computed and applied to all frames. The aligned sequences are stored separately as TrainSet2, while the original sequences are retained. The same alignment procedure is applied to the gallery and probe sets during testing. The model is based on the 3D variant of DeepGaitV2 provided by the OpenGait framework. Given the larger size of TrainSet1, training is extended to 120,000 iterations, with learning rate decay scheduled at 60k, 90k, and 110k steps. A probabilistic sampling strategy is used, where identities are drawn proportionally from each dataset, and sequences are selected within the same domain to reduce variance.

Four models are trained using forward and reversed sequences from both TrainSet1 and TrainSet2. During inference, their distance matrices are averaged and re-ranked, followed by final refinement using a Hungarian-based capacity-constrained matching algorithm~\cite{kuhn1955hungarian}.

\vspace{-0.5em}
\subsection{Team: league}
\vspace{-0.5em}

\noindent \textbf{Member:} Xue Cheng, Lichen Song, and Junjian Cui \\
\noindent \textbf{Affiliation:} EVERSPRY \\
\noindent \textbf{Method:} 
The training data comprises samples from two public datasets, Gait3D and HID2022, and a proprietary dataset collected in both real-world and controlled environments. Pseudo-labels for HID2022 test data are generated using the winning model from the HID2022 challenge. The internal dataset consists of surveillance footage around office premises and laboratory-collected sequences, which are not publicly released due to compliance requirements.

All data undergo PCA-based rotational alignment, where the principal axis of each silhouette is estimated and aligned vertically to ensure a consistent upright posture. Identity classes with fewer than five sequences are excluded to maintain sufficient intra-class diversity. Data augmentation includes random affine transformations, horizontal flipping, and random rotations applied during training to improve generalization. Two backbone models are employed: DeepGaitV2 and SwinGait3D. The former adopts a 22-layer ResNet with Pseudo-3D convolution modules for compact spatial-temporal representation learning, while the latter incorporates 3D Swin Transformer blocks with window-based attention to model long-range dependencies. Both models follow a similar pipeline, differing mainly in feature encoding architecture.

\begin{figure}[htbp]
	\centering
    \vspace{-1.0em}
	\includegraphics[width=\linewidth]{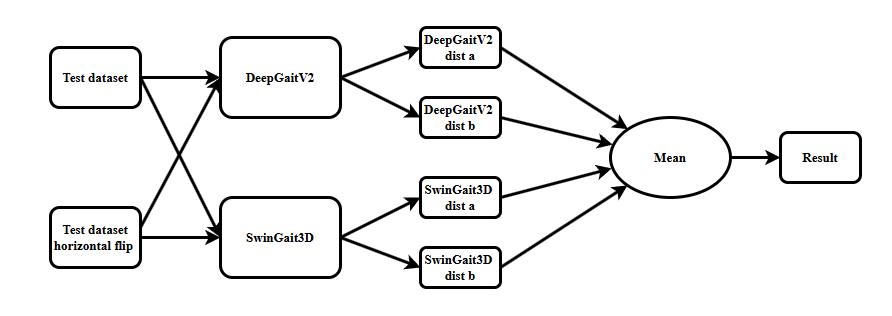}
    \vspace{-1.0em}
	\caption{The multi-model fusion strategy of Team league's method.}
    \vspace{-1.0em}
	\label{fig:team5_strategy}
\end{figure}

During inference, a multi-model ensemble strategy is applied, as shown in Figure~\ref{fig:team5_strategy}. Each model processes the original and horizontally flipped versions of the test set. For each probe-gallery pair, distance matrices from all inference paths are averaged to compute the final similarity scores. To enhance final prediction quality, a capacity-constrained matching algorithm based on the Hungarian method is applied. This pipeline improves recognition robustness by integrating predictions from diverse model types and test augmentations.

\vspace{-0.5em}
\subsection{Team: HUST-MCLAB}
\vspace{-0.5em}

\noindent \textbf{Members:} Junjie Zhou (Wuhan University of Technology), Haijun Xiong (Huazhong University of Science and Technology), and Junhao Lu (Hefei University of Technology) \\
\noindent \textbf{Supervisor:} Bin Feng (Huazhong University of Science and Technology) \\
\noindent{\small \texttt{zhoujunjie@whut.edu.cn, \{xionghj, fengbin\}@hust.edu.cn, 2021214872@mail.hfut.edu.cn}}

\begin{figure}[htbp]
    \centering
    \vspace{-1.0em}
    \includegraphics[width=0.8\linewidth]{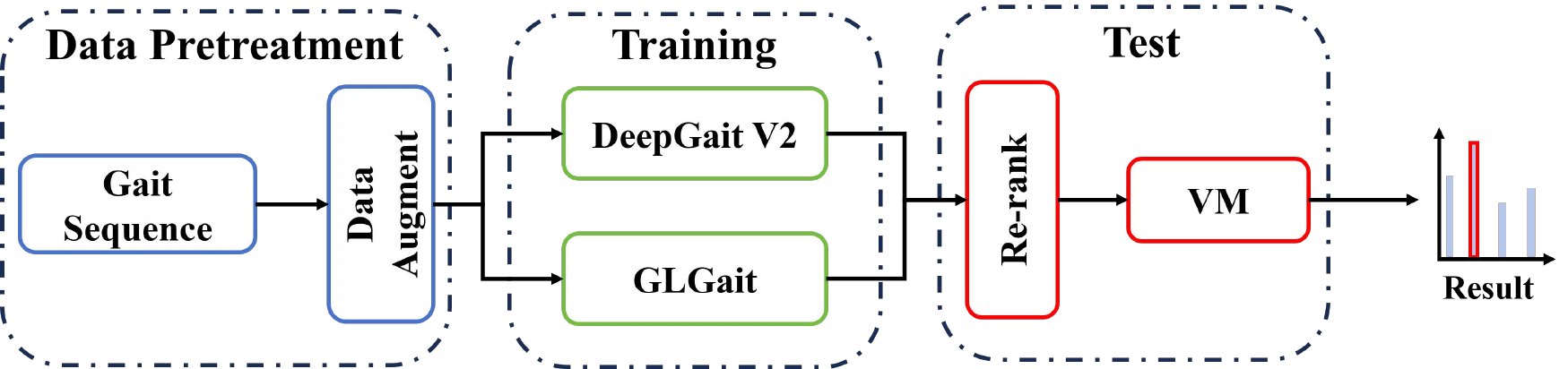}
    \caption{The framework of Team HUST-MCLAB’s method.}
    \vspace{-1.0em}
    \label{fig:6.1}
\end{figure}

\noindent\textbf{Method:} As shown in Figure~\ref{fig:6.1}, this solution contains three components: data pretreatment, training, and testing.

\noindent\textbf{(1) Data Pretreatment:} For a given set of gait sequences, the first preprocessing step is to resize them to a resolution of 64×64 pixels. Subsequently, three common data augmentation strategies, including random perspective transformation, random horizontal flipping, and random rotation, are employed to enhance dataset variability and model generalization.

\noindent\textbf{(2) Training:} In this method, two models, DeepGait V2 (P3D)~\cite{fan2023exploring} and GLGait~\cite{peng2024glgait}, are employed. Both models are trained on the SUSTech1K~\cite{lidargait}, Gait3D~\cite{gait3d_cvpr2022} and HID 2022 datasets. Specifically, for GLGait, the Cosine and Euclidean distances are summed to form the overall distance metric used in the triplet loss to better distinguish samples that are similar but not identical.

\noindent\textbf{(3) Test:} During the test phase, Re-ranking (RK) and Vote Mechanism (VM) are utilized to further improve the accuracy of the results. RK is applied  to strengthen the feature representation, while VM aggregates the capabilities of multiple models to improve the final accuracy. In particular, DeepGait V2 (P3D) and GLGait trained for different iterations are aggregated through a VM. The individual model outputs and the final fused results are presented in the Table~\ref{tab:model_accuracy}. It can be observed that VM effectively improves the overall recognition accuracy.

\begin{table}[htbp]
\centering
\vspace{-1.0em}
\caption{The rank - 1 accuracies by different models}
\label{tab:model_accuracy}
\resizebox{0.35\textwidth}{!}{%
\begin{tabular}{c | c | c}
\hline
Model & Iteration & Accuracy (\%) \\
\hline
DeepGait V2 (P3D) & 110000  & 87.8  \\
DeepGait V2 (P3D) & 120000  & 87.6  \\
GLGait & 100000  & 86.6  \\
GLGait & 110000  & 87.2  \\
GLGait & 120000  & 87.8  \\
VM & - & 88.2 \\
\hline
\end{tabular}
}
\vspace{-1.0em}
\end{table}

\subsection{Team: ITZY}
\vspace{-0.5em}

\noindent \textbf{Members:} Yanfei Wang (School of Intelligent Engineering, Sun Yat-sen University) \\
\noindent \textbf{Supervisor:} Mengyuan Liu (Peking University, Shenzhen Graduate School) \\
\noindent{\small \texttt{Wangyf387@mail2.sysu.edu.cn, liumengyuan@pku.edu.cn}}

\begin{figure}[htbp]
    \centering
    \vspace{-1.0em}
    \includegraphics[width=1\linewidth]{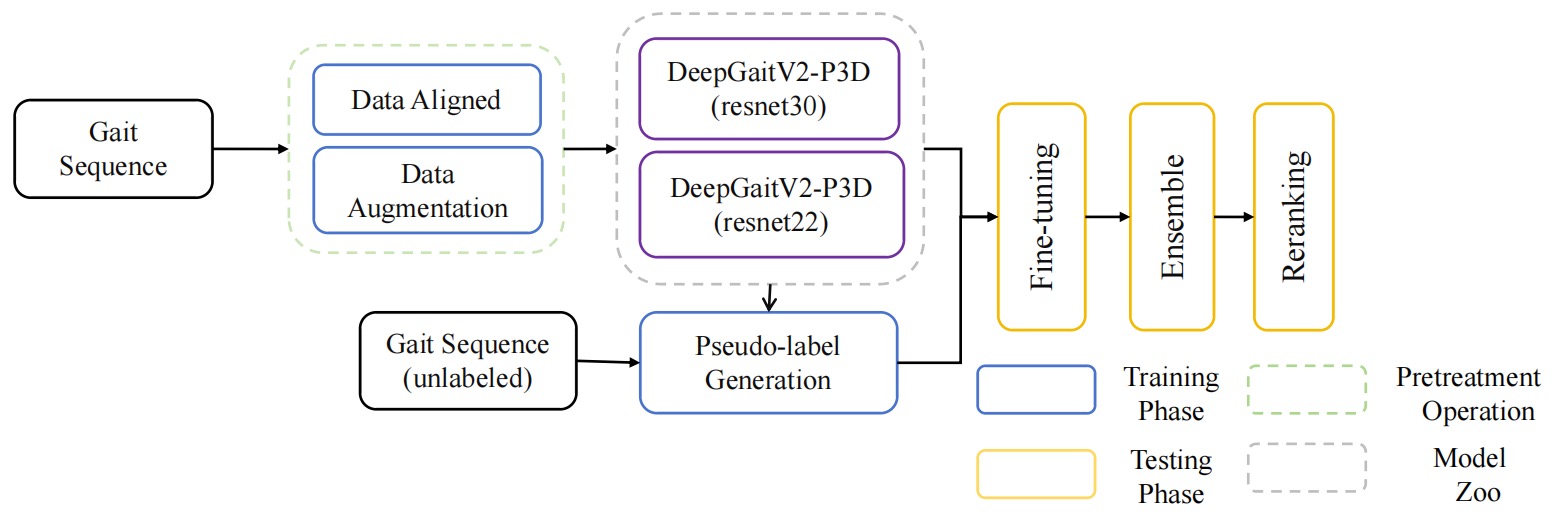}
    \vspace{-1.0em}
    \caption{The framework of Team ITZY’s method.}
    \vspace{-1.0em}
    \label{fig:7.1}
\end{figure}

\noindent \textbf{Method:} 
This method proposes a three-stage gait recognition framework that integrates multi-dataset pretraining, pseudo-label refinement, and ensemble-based inference, as depicted in Figure~\ref{fig:7.1}. 

The training data combines silhouettes from CASIA-E~\cite{casiae}, Gait3D~\cite{gait3d_cvpr2022}, CASIA-B~\cite{casiab}, CCPG~\cite{ccpg}, SUSTech1K~\cite{lidargait}, and selected sequences from OU-MVLP~\cite{dataset2017OUMVLP}. All sequences are aligned using PCA-based rotation, resizing to $64 \times 64$, and augmented via horizontal flipping, random rotation, and perspective transformation with respective probabilities of 0.3, 0.4, and 0.4.

The training follows a three-phase pipeline. In the first phase, DeepGaitV2-P3D~\cite{fan2023exploring} models with ResNet-22 and ResNet-30 backbones are pretrained for 200,000 iterations on the combined dataset using the OpenGait framework. In the second phase, these models are used to generate pseudo-labels for the HID2022~\cite{hid2022summary} test set, which contains 504 subjects. In the third phase, fine-tuning is conducted on a mixed dataset of original and pseudo-labeled samples, where only the final two layers of the backbone are updated using triplet loss.

Inference is based on a weighted ensemble of both models. For each probe-gallery pair, distances are computed across models and fused as: $y_{\text{final}} = \text{argsort}\left(\sum_{i=1}^{2} w_i f_{\theta_i}(x)\right),$
where $w = [0.6, 0.4]$ represent weights for ResNet-22 and ResNet-30, respectively. A re-ranking mechanism is applied to the averaged distance matrix to further improve recognition performance.

Experimental results on the HID2025 validation set show that multi-dataset pretraining improves rank-1 accuracy from 45.9\% to 81.5\%. Incorporating pseudo-label fine-tuning increases accuracy to 86.7\%, and ensemble voting leads to a final score of 86.9\%.

\vspace{-0.5em}
\subsection{Team: sysu}
\vspace{-0.5em}

\noindent \textbf{Members:} Li Jiahang and Zhou Ji \\
\noindent \textbf{Supervisor:} Zhao Baoquan \\
\noindent \textbf{Institutions:} Sun Yat-sen University \\
\noindent{\small \texttt{\{lijh396, zhouj527, zhaobaoquan\}@mail.sysu.edu.cn}}

\noindent \textbf{Method:} 
This method introduces a two-stage gait recognition pipeline based on a modified DeepGaitV2 architecture, trained on a merged multi-dataset corpus and fine-tuned on task-specific data with re-ranking refinement. 

The training data consists of four benchmark datasets—CASIA-B, CCPG, Gait3D, and SUSTech1K—containing a total of 5,104 subjects. To ensure data consistency, standardized preprocessing procedures are applied, including silhouette normalization and spatial alignment across all sequences, enabling stable cross-dataset training and feature extraction.

The model architecture is derived from DeepGaitV2, with modifications to the backbone structure in a 2-6-4-1 layer configuration to improve discriminative capacity. Each training batch includes 20 subjects, with six sequences per subject and 15 frames per sequence. The model is trained for 100,000 iterations using the default hyperparameter settings of the DeepGaitV2 framework.

The training pipeline comprises two phases. In the first phase, the model is trained on the merged dataset to obtain generalizable weights. In the second phase, fine-tuning is conducted using the gallery subset of the HID dataset to adapt to domain-specific features. A conservative learning rate of $10^{-5}$ and a brief training schedule of 500 iterations are used to prevent overfitting due to limited data volume. To further enhance retrieval accuracy, a re-ranking strategy is enabled during evaluation, which leads to noticeable improvements in recognition performance.

\vspace{-0.5em}
\subsection{Team: SJTU-ICL}
\vspace{-0.5em}

\noindent \textbf{Members:} Zepeng Wang and Ke Xu \\
\noindent \textbf{Institutions:} Shanghai Jiao Tong University \\
\noindent{\small \texttt{\{wzp.ck, l13025816\}@sjtu.edu.cn}}

\noindent \textbf{Method:}
This method uses DeepGaitV2 (CNN-based) and SwinGait (Transformer-based) for gait recognition. Training is conducted on a merged dataset comprising CASIA-E~\cite{casiae}, CCPG~\cite{ccpg}, CASIA-B~\cite{casiab}, Gait3D~\cite{gait3d_cvpr2022}, SUSTech1K~\cite{lidargait}, OU-MVLP~\cite{dataset2017OUMVLP}, and GREW~\cite{grew_iccv2021}. To reduce forgetting, datasets are trained sequentially before joint optimization. All inputs are resized to $128\times128$, except OU-MVLP and GREW which are upsampled from $64\times64$; the final input size is $128\times88$ after BaseSilCuttingTransform.

Augmentations include flipping, rotation, and random erasing with a 0.5 probability. Six model variants are trained with different dataset and augmentation settings. Cosine similarity is used during testing, improving accuracy by ~0.5\%. Final predictions are determined by majority voting; if predictions differ, the result from the most accurate model is selected. Class distribution across five-query samples is analyzed to assess submission quality, following Phase-2’s 90\% dominance rule.

Experiments are run on four A100 GPUs using OpenGait~\cite{fan2023exploring}. Initial training uses SGD for 360k iterations with learning rate decay at 160k, 240k, and 300k. GaitMix is applied in early training. Fine-tuning uses AdamW with a learning rate decay from 1e-4 to 1e-5 at 160k. Batch size is $32\times4$, representing 32 identities and 4 samples per identity.

\vspace{-0.5em}
\section{Analysis}
\vspace{-0.5em}
The technical solutions of the top HID 2025 teams are summarized in Table~\ref{tab:top2025}. Data cleaning, alignment, and augmentation were standard steps for handling noise and improving sample consistency. All top teams combined multiple public datasets to increase data diversity and robustness. Ensemble methods and re-ranking were commonly used to improve recognition accuracy, including model fusion and checkpoint voting strategies. Several teams used pseudo-labelling to leverage unlabelled or weakly labelled data.

Compared with previous years, \textbf{HID 2025 had two main differences in technical approaches}. \textbf{First}, the top two teams fine-tuned their models using the HID 2025 gallery samples. This result shows that fine-tuning a strong pre-trained backbone with only a small amount of target data can yield notable gains in cross-domain recognition accuracy. Such a strategy is practical for real-world deployment in new environments, for example, when applying gait recognition to a new surveillance site where only a few labelled samples are available. Fine-tuning with this small set enables rapid and robust adaptation to the new domain without the need for collecting and annotating a large local dataset. \textbf{Second}, more teams focused on spatial alignment of the gait sequences. The goal of alignment is to standardize the walking posture, making the subjects appear upright in the silhouette, similar to the concept of facial alignment in face recognition.
Most teams applied unsupervised alignment methods such as PCA-based rotation. The winning team further used skeleton-based alignment, which provided higher alignment accuracy and directly improved the final recognition performance.

\textbf{Another notable observation} is that all top teams adopted DeepGaitV2~\cite{fan2023exploring} as their backbones, reflecting the current consensus on the importance of strong model architectures in gait recognition. However, the further progress of backbone models depends on larger and more diverse datasets, which are difficult to collect and annotate in gait research. Some recent studies have shown that large vision models trained on general images or videos can be transferred to gait recognition tasks~\cite{gaitprompt24,biggait}. This approach may help address the data limitation problem and improve gait recognition in the future.

\vspace{-0.5em}
\section{Conclusions and Future Paths}

The 6th International Competition on Human Identification at a Distance (HID 2025) provided a comprehensive evaluation of gait recognition methods under challenging, real-world conditions. 
The competition attracted over 113 registered teams, which reflects both the growing interest and technical maturity in the field.
This year’s competition maintained the use of the SUSTech-Competition dataset, which includes a wide range of variations in clothing, carried objects, and viewpoints. 
No official training set was provided, so all participants trained their models on external data. 
Under these conditions, several teams achieved new benchmarks, with the best accuracy reaching 94.2\%. 
This result shows clear progress in the robustness and generalization of gait recognition methods.

One important finding from HID 2025 is that spatial alignment and domain adaptation approaches can achieve strong performance. Although some teams used complex ensembles or many additional techniques, most top results were obtained by refining existing deep learning backbones and focusing on data quality, training strategies, and model adaptation. This suggests that future research should continue to improve core model capability, rather than relying on increasing complexity.

While HID 2025 successfully benchmarked the state of the art in gait recognition, a critical reflection on its design reveals opportunities for future improvement that can better guide the field. First, the evaluation relied solely on Rank-1 accuracy. While straightforward, this single metric may not fully capture a model's overall identification capabilities, especially its performance in less-than-ideal matches. Second, as insightful reviewers noted, the competition dataset—though challenging—lacks the extreme "in-the-wild" scenarios such as adverse weather, severe occlusions from crowds, or low-resolution footage. Therefore, future HID competitions could evolve in two key ways: by adopting a more comprehensive evaluation suite, such as including Rank-5 accuracy and the Area Under the CMC curve (AUCMC), and by introducing new datasets or special tracks that specifically target these unsolved "in-the-wild" conditions to incentivize research in these critical areas.

Building upon the successes and insights from HID 2025, the path forward for gait recognition research is now clearer, focusing on several promising directions to tackle remaining challenges. (1) \textbf{Cross-domain generalization with large pretrained models}. Using large vision models pretrained on large and diverse datasets can help gait recognition methods adapt to new domains and tasks with limited labeled data~\cite{biggait, jin2025denoising}. Leveraging these foundation models could improve generalization in scenarios with domain shift. (2) \textbf{Multi-modal integration with diverse data sources}. Integrating data sources such as 3D structure, skeletal information, or scene context with silhouette features may help address specific challenges in gait recognition, including occlusions, extreme viewpoint variations, and complex backgrounds. This also presents a prime opportunity to innovate beyond the current architectural consensus, where nearly all top teams used DeepGaitV2. Future work could focus on developing novel fusion strategies that leverage Transformer-based architectures~\cite{dosovitskiy2020image}, which are inherently adept at integrating heterogeneous data streams. While some teams in HID 2025 already explored this path with models like SwinGait~\cite{fan2023exploring}, significant research is needed to unlock their full potential for robust multi-modal recognition. (3) \textbf{Privacy-preserving synthetic data generation with video generation models}. Considering privacy issues, collecting large-scale and diverse gait datasets—especially those with multiple modalities—is a significant challenge. Recent progress in video generation techniques, such as Sora~\cite{openai_sora_2024}, Veo 3~\cite{deepmind_veo3_2025}, and KelingAI~\cite{klingai_2024}, provides an alternative by enabling the creation of synthetic gait samples that can supplement real data. Leveraging advanced generative models may help address data scarcity, support training under privacy constraints~\cite{ma2024passersby}, and promote further research in gait recognition without the need for extensive real-world data collection. Ultimately, these paths hold promise to propel gait recognition beyond controlled environments and into the complexities of the real world, thereby fulfilling its potential as a secure and practical identification modality.

\vspace{-0.5em}
\section*{Acknowledgements}
We express our sincere appreciation to Watrix Technology Limited Co. Ltd. for their generous sponsorship of the competition, which played a crucial role in its success. We would also like to express our gratitude to CodaLab for providing the result evaluation platform, enabling a seamless and efficient evaluation process. We are deeply grateful to our advisors, Prof. Mark Nixon, Prof. Tieniu Tan, and Prof. Yasushi Yagi, for their invaluable support and insightful suggestions throughout the competition. We would also like to acknowledge the technical support provided by Mr. Rui Wang. Their assistance and expertise greatly contributed to the successful organization and execution of the competition.

This work was supported by the National Natural Science Foundation of China (Grant No. 62476120), the Key Research Project of ShenZhen Polytechnic University (No. 6025310010K), and the Shenzhen Polytechnic University Research Fund (No. 6025310044K).

{\small
\bibliographystyle{ieee}
\bibliography{egbib}

@INPROCEEDINGS{hid2022summary,
  author={Yu, Shiqi and Huang, Yongzhen and Wang, Liang and Makihara, Yasushi and Wang, Shengjin and Rahman Ahad, Md Atiqur and Nixon, Mark},
  booktitle={IEEE International Joint Conference on Biometrics (IJCB)}, 
  title={{HID 2022}: The 3rd International Competition on Human Identification at a Distance}, 
  year={2022},
  pages={1-9},
}

@ARTICLE{casiae,
  author={Song, Chunfeng and Huang, Yongzhen and Wang, Weining and Wang, Liang},
  journal={IEEE Transactions on Pattern Analysis and Machine Intelligence}, 
  title={{CASIA-E}: A Large Comprehensive Dataset for Gait Recognition}, 
  year={2023},
  volume={45},
  number={3},
  pages={2801-2815}
}

@article{dataset2017oumvlp,
 author = {Noriko Takemura and Yasushi Makihara and Daigo Muramatsu and Tomio Echigo and Yasushi Yagi},
 journal = {IPSJ Transactions on Computer Vision and Applications},
 number = {4},
 pages = {1-14},
 title = {Multi-view large population gait dataset and its performance evaluation for cross-view gait recognition},
 volume = {10},
 year = {2018}
}

@ARTICLE{gaitset,  
    author={Chao, Hanqing and Wang, Kun and He, Yiwei and Zhang, Junping and Feng, Jianfeng},  
    journal={IEEE Transactions on Pattern Analysis and Machine Intelligence},   
    title={{GaitSet}: Cross-View Gait Recognition Through Utilizing Gait As a Deep Set},   
    year={2022},  
    volume={44},  
    number={7},  
    pages={3467-3478},  
}

@inproceedings {grew_iccv2021,
title=  {Gait Recognition in the Wild: A Benchmark},
author=  {Zheng Zhu and Xianda Guo and Tian Yang and Junjie Huang and Jiankang Deng and Guan Huang and Dalong Du and Jiwen Lu and Jie Zhou},
booktitle=  {IEEE/CVF International Conference on Computer Vision (ICCV)},
year=  {2021}              
}

@inproceedings{gait3d_cvpr2022,
title={Gait Recognition in the Wild with Dense 3D Representations and A Benchmark},
author={Jinkai Zheng and Xinchen Liu and Wu Liu, Lingxiao He and Chenggang Yan and Tao Mei},
booktitle={IEEE/CVF Conference on Computer Vision and Pattern Recognition (CVPR)},
year={2022}
}

@inproceedings{casiab,
 author = {Yu, Shiqi and Tan, Daoliang and Tan, Tieniu},
 booktitle = {International Conference on Pattern Recognition (ICPR)},
 publisher = {IEEE},
 pages = {441--444},
 title = {A framework for evaluating the effect of view angle, clothing and carrying condition on gait recognition},
 volume = {4},
 year = {2006}
}

@InProceedings{biggait,
    author    = {Ye, Dingqiang and Fan, Chao and Ma, Jingzhe and Liu, Xiaoming and Yu, Shiqi},
    title     = {{BigGait}: Learning Gait Representation You Want by Large Vision Models},
    booktitle = {IEEE/CVF Conference on Computer Vision and Pattern Recognition (CVPR)},
    month     = {June},
    year      = {2024},
    pages     = {200-210}
}

@InProceedings{ccpg,
    author    = {Li, Weijia and Hou, Saihui and Zhang, Chunjie and Cao, Chunshui and Liu, Xu and Huang, Yongzhen and Zhao, Yao},
    title     = {An In-Depth Exploration of Person Re-Identification and Gait Recognition in Cloth-Changing Conditions},
    booktitle = {IEEE/CVF Conference on Computer Vision and Pattern Recognition (CVPR)},
    month     = {June},
    year      = {2023},
    pages     = {13824-13833}
}

@inproceedings{ccgr,
  title={Cross-Covariate Gait Recognition: A Benchmark},
  author={Zou, Shinan and Fan, Chao and Xiong, Jianbo and Shen, Chuanfu and Yu, Shiqi and Tang, Jin},
  booktitle={AAAI Conference on Artificial Intelligence (AAAI)},
  year={2024}
}

@InProceedings{lidargait,
    author    = {Shen, Chuanfu and Fan, Chao and Wu, Wei and Wang, Rui and Huang, George Q. and Yu, Shiqi},
    title     = {LidarGait: Benchmarking 3D Gait Recognition With Point Clouds},
    booktitle = {IEEE/CVF Conference on Computer Vision and Pattern Recognition (CVPR)},
    month     = {June},
    year      = {2023},
    pages     = {1054-1063}
}

@InProceedings{gaitprompt24,
    author    = {Ma, Kang and Fu, Ying and Cao, Chunshui and Hou, Saihui and Huang, Yongzhen and Zheng, Dezhi},
    title     = {Learning Visual Prompt for Gait Recognition},
    booktitle = {IEEE/CVF Conference on Computer Vision and Pattern Recognition (CVPR)},
    month     = {June},
    year      = {2024},
    pages     = {593-603}
}

@ARTICLE{drone24,
  author={Li, Aoqi and Hou, Saihui and Cai, Qingyuan and Fu, Yang and Huang, Yongzhen},
  journal={IEEE Transactions on Multimedia}, 
  title={Gait Recognition With Drones: A Benchmark}, 
  year={2024},
  volume={26},
  pages={3530-3540},
  doi={10.1109/TMM.2023.3312931}
}

@ARTICLE{problem2005,
  author={Sarkar, S. and Phillips, P.J. and Liu, Z. and Vega, I.R. and Grother, P. and Bowyer, K.W.},
  journal={IEEE Transactions on Pattern Analysis and Machine Intelligence}, 
  title={The humanID gait challenge problem: data sets, performance, and analysis}, 
  year={2005},
  volume={27},
  number={2},
  pages={162-177},
  doi={10.1109/TPAMI.2005.39}
}

@book{nixon2006book,
    author = {Mark S. Nixon and Tieniu Tan and Rama Chellappa},
    title = {Human Identification Based on Gait},
    publisher = {Springer},
    year = 2006
}

@article{connor2018,
author = {Patrick Connor and Arun Ross},
title = {Biometric recognition by gait: A survey of modalities and features},
journal = {Computer Vision and Image Understanding},
volume = {167},
pages = {1-27},
year = {2018},
}

@inproceedings{jin2025denoising,
  title={On Denoising Walking Videos for Gait Recognition},
  author={Jin, Dongyang and Fan, Chao and Ma, Jingzhe and Zhou, Jingkai and Chen, Weihua and Yu, Shiqi},
  booktitle={Proceedings of the Computer Vision and Pattern Recognition Conference},
  pages={12347--12357},
  year={2025}
}

@inproceedings{peng2024glgait,
  title={Glgait: a global-local temporal receptive field network for gait recognition in the wild},
  author={Peng, Guozhen and Wang, Yunhong and Zhao, Yuwei and Zhang, Shaoxiong and Li, Annan},
  booktitle={Proceedings of the 32nd ACM International Conference on Multimedia},
  pages={826--835},
  year={2024}
}

@article{kuhn1955hungarian,
  title={The Hungarian method for the assignment problem},
  author={Kuhn, Harold W},
  journal={Naval research logistics quarterly},
  volume={2},
  number={1-2},
  pages={83--97},
  year={1955},
  publisher={Wiley Online Library}
}

@misc{openai_sora_2024,
  author       = {OpenAI},
  title        = {Sora System Card},
  howpublished = {\url{https://openai.com/index/sora-system-card/}}
}

@misc{deepmind_veo3_2025,
  author       = {Google DeepMind},
  title        = {Veo 3},
  howpublished = {\url{https://deepmind.google/models/veo/}}
}

@misc{klingai_2024,
  author       = {Kuaishou Technology},
  title        = {Kling AI},
  howpublished = {\url{https://kling-ai.video/}}
}

@article{dosovitskiy2020image,
  title={An image is worth 16x16 words: Transformers for image recognition at scale},
  author={Dosovitskiy, Alexey and Beyer, Lucas and Kolesnikov, Alexander and Weissenborn, Dirk and Zhai, Xiaohua and Unterthiner, Thomas and Dehghani, Mostafa and Minderer, Matthias and Heigold, Georg and Gelly, Sylvain and others},
  journal={arXiv preprint arXiv:2010.11929},
  year={2020}
}

@article{fan2023exploring,
  title={OpenGait: A Comprehensive Benchmark Study for Gait Recognition towards Better Practicality},
  author={Fan, Chao and Hou, Saihui and Liang, Junhao and Shen, Chuanfu and Ma, Jingzhe and Jin, Dongyang and Huang, Yongzhen and Yu, Shiqi},
  journal={IEEE Transactions on Pattern Analysis and Machine Intelligence},
  year={2025},
  publisher={IEEE}
}

@inproceedings{ma2024passersby,
  title={Passersby-Anonymizer: Safeguard the Privacy of Passersby in Social Videos},
  author={Ma, Jingzhe and Luo, Haoyu and Huang, Zixu and Jin, Dongyang and Wang, Rui and Briffa, Johann A and Poh, Norman and Yu, Shiqi},
  booktitle={2024 IEEE International Joint Conference on Biometrics (IJCB)},
  pages={1--10},
  year={2024},
  organization={IEEE}
}
}

\end{document}